\documentclass[12pt]{article}
\usepackage[latin1]{inputenc}
\usepackage[margin=1.22in]{geometry}
\usepackage{url}
\usepackage{natbib}
\usepackage{latexsym}
\usepackage{url}
\usepackage{graphicx}
\usepackage{xcolor}
\usepackage{longtable}

\newcommand{\CLCOMMENT}[1]{\multicolumn{2}{p{0.9\linewidth}}{\textcolor{gray}{#1}}\\}
\newcommand{\CELAPSE}{\multicolumn{2}{c}{\ldots}\\}

\title{A Roadmap towards Machine Intelligence}

\author{Tomas Mikolov$^1$, Armand Joulin$^1$ and Marco Baroni$^{1,2}$\\\\
{\small $^1$Facebook AI Research}\\
{\small $^2$University of Trento}}

\date{}

\begin{document}
\maketitle

\begin{abstract}
  The development of intelligent machines is one of the biggest
  unsolved challenges in computer science. In this paper, we propose
  some fundamental properties these machines should have, focusing in
  particular on \emph{communication} and \emph{learning}. We discuss a
  simple environment that could be used to incrementally teach a
  machine 
  the basics of natural-language-based communication, as a
  prerequisite to 
  more complex interaction with human users. We also present some
  conjectures on the sort of algorithms the machine should support in
  order to profitably learn from the environment.
\end{abstract}

\section{Introduction}
\label{sec:introduction}




A machine capable of performing complex tasks without requiring
laborious programming would be tremendously useful in almost any human
endeavor, from performing menial jobs for us to helping the
advancement of basic %
and applied research.  
Given the current availability of powerful hardware and large amounts
of machine-readable data, as well as the widespread interest in
sophisticated machine learning methods, the times should be ripe for
the development of intelligent machines. 

Still, since ``solving
AI'' seems too complex a task to be pursued all at once, in the last decades the
computational community has preferred to focus
on solving relatively narrow empirical problems that are important for
specific applications, but do not address the overarching goal of
developing general-purpose intelligent machines. In this article, we
propose an alternative approach: we first define the general
characteristics we think intelligent machines should possess, and then
we present a concrete roadmap to develop them in realistic, small
steps, that are however incrementally structured in such a way that,
jointly, they should lead us close to the ultimate goal of
implementing a powerful AI.



The article is organized as follows. In Section \ref{sec:desiderata}
we specify the two fundamental characteristics that we consider
crucial for developing intelligence--at least the sort of intelligence
we are interested in--namely \emph{communication} and
\emph{learning}. Our goal is to build a machine that can learn new
concepts through communication at a similar rate as a human with
similar prior knowledge. That is, if one can easily learn how
subtraction works after mastering addition, the intelligent machine,
after grasping the concept of addition, should not find it difficult
to learn subtraction as well. Since, as we said, achieving the long-term goal of building an
intelligent machine equipped with the desired features at once seems
too difficult, we need to define intermediate targets that can lead us
in the right direction. %
We specify such targets in terms of simplified but self-contained
versions of the final machine we want to develop. At any time during its
``education'', the target machine should act like a stand-alone
intelligent system, albeit one that will be initially very limited in
what it can do. The bulk of our proposal (Section \ref{sec:thegame})
thus consists in the plan for an interactive learning environment
fostering the incremental development of progressively more
intelligent behavior. Section \ref{sec:computational-systems} briefly discusses some of the
algorithmic capabilities we think a machine should possess in order to
profitably exploit the learning environment. Finally, Section
\ref{sec:related} situates our proposal in the broader context of past
and current attempts to develop intelligent machines. As that review
should make clear, our plan encompasses many ideas that have already
appeared in different research strands. What we believe to be novel in our approach is the way in which we are combining such ideas into a coherent program.

\section{Desiderata for an intelligent machine}
\label{sec:desiderata}

Rather than attempting to formally characterize intelligence, we
propose here a set of desiderata we believe to be crucial for a
machine to be able to autonomously make itself helpful to humans in
their endeavors. The guiding principles we implicitly considered in
formulating the desiderata are to minimize the complexity of the
machine, and to maximize interpretability of its behavior by humans.






\subsection{Ability to communicate} %
\label{sec:communication}
Any practical realization of an intelligent machine will have to
\emph{communicate} with us. It would be senseless to build a machine
that is supposed to perform complex operations if there is no way for
us to specify the aims of these operations, or to understand the
output of the machine. While other communication means could be
entertained, natural language is by far the easiest and most powerful
communication device we possess, so it is reasonable to require an
intelligent machine to be able to communicate through
language. Indeed, the intelligent machine we aim for could be seen as
a computer that can be programmed through natural language, or as the
interface between natural language and a traditional programming
language.  Importantly, humans have encoded a very large portion of
their knowledge into natural language (ranging from mathematics
treatises to cooking books), so a system mastering natural language
will have access to most of the knowledge humans have assembled over
the course of their history.

Communication is, by its very nature, \emph{interactive}: the
possibility to hold a conversation is crucial both to gather new
information (asking for explanation, clarification, instructions,
feedback, etc.)~and to optimize its transmission (compare a good
lecture or studying with a group of peers to reading a book
alone). Our learning environment will thus emphasize the interactive
nature of communication.

Natural language can also channel, to a certain extent, non-linguistic
information, because much of the latter can be conveyed through
linguistic means. For example, we can use language to talk about what
we perceive with our senses, or to give instructions on how to operate
in the world (see \citealp{Louwerse:2011}, among others, for evidence that language encodes
many perceptual aspects of our knowledge). Analogously, in the simulation
we discuss below, a Teacher uses natural language to teach the Learner
(the intelligent machine being trained) a more limited and explicit
language (not unlike a simple programming language) in which the
Learner can issue instructions to its environment through the same
communication channels it uses to interact with the
Teacher. 
The intelligent machine can later be instructed to browse the
Internet by issuing commands in the appropriate code through its usual
communication channels, mastering in this way a powerful tool to
interact with the world at large. Language can also serve as an
interface to perceptual components, and thus update the machine about
its physical surroundings. For example, an object recognition system
could transform raw pixel data into object labels, allowing the
machine to ``see'' its real-life environment through a
controlled-language modality.

Still, we realize that our focus on the language-mediated side of
intelligence may limit the learning machine in the development of
skills that we naturally gain by observing the world around us. There
seems to be a fundamental difference between the symbolic
representations of language and the continuous nature of the world as we perceive it. If this will turn out to be an issue, we
can extend the training phase of the machine (its development in a
simulated environment such as the one we will sketch below) with tasks
that are more perception-oriented. While in the tasks we will describe
here the machine will be taught how to use its I/O channels to
receive and transmit linguistic symbols, the machine could also be exposed,
through the same interface,
to simple encodings (bit streams) of continuous input
signals, such as images.  The machine could thus be trained, first, to
understand the basic properties of continuous variables, and then to
perform more complex operations in a continuous space, such as
identifying shapes in 2D images. Note that including such tasks would
not require us to change the design of our learning framework, only to
introduce novel scripts.

One big advantage of the single-interface approach we are currently
pursuing is that the machine only needs to be equipped with bit-based
I/O channels, thus being maximally simple in its interface. The
machine can learn an unlimited number of new codes enabling it to
interface, through the same channels, with all sorts of interlocutors
(people, other machines, perceptual data encoded as described above,
etc.). 
By equipping the machine with only a minimalistic I/O bit-stream
interface, we ensure moreover that no prior knowledge about the
challenges the machine will encounter is encoded into the structure of
the input and output representations, harming the generality of the
strategies the machine will learn (compare the difficulty of
processing an image when it's already encoded into pixels vs.~as raw
bits).

Finally, while we propose language as the general \emph{interface} to
the machine, we are agnostic about the nature of the internal
representations the machine must posit to deal with the challenges it
faces. In particular, we are not making claims about the internal
representations of the machine being based on an interpretable
``language of thought'' \citep{Fodor:1975}. In other words, we are not
claiming that the machine should carry out its internal reasoning in a
linguistic form: only that its input and output are linguistic in
nature.

To give a few examples of how a communication-based intelligent
machine can be useful, consider a  machine helping a
scientist with research. First of all, the communication-endowed
machine does not need to pre-encode a large static database of facts,
since it can retrieve the relevant information from the Internet. If
the scientist asks a simple question such as: \emph{What is the
  density of gold?}, the machine can search the Web to answer:
\emph{$19.3 g/cm^3$}. 

Most questions will however require the machine to put together
multiple sources of information. For example, one may ask: \emph{What
  is a good starting point to study reinforcement learning?}. The
machine might visit multiple Web sites to search for materials and get
an idea of their relative popularity. Moreover, interaction can make
even a relatively simple query such as the latter more successful. For
example, the machine can ask the user if she prefers videos or
articles, what is the mathematical background to be assumed, etc.

However, what we are really interested in is a machine that can
significantly speed up research progress by being able to address
questions such as: \emph{What is the most promising direction to cure
  cancer, and where should I start to meaningfully contribute?} This
question may be answered after the machine reads a significant number
of research articles online, while keeping in mind the perspective of
the person asking the question. Interaction will again play a central
role, as the best course of action for the intelligent machine might
involve entering a conversation with the requester, to understand her
motivation, skills, the time she is willing to spend on the topic,
etc. Going further, in order to fulfill the request above, the machine
might even conduct some independent research by exploiting information
available online, possibly consult with experts, and direct the
budding researcher, through multiple interactive sessions, towards
accomplishing her goal.

\subsection{Ability to learn}

Arguably, the main flaw of ``good old'' symbolic AI research
\citep{Haugeland:1985} lied in the assumption that it would be
possible to program an intelligent machine largely by hand. We believe
it is uncontroversial that a machine supposed to be helping us in a
variety of scenarios, many unforeseen by its developers, should be
endowed with the capability of \emph{learning}. A machine that does
not learn cannot adapt or modify itself based on experience, as it
will react in the same way to a given situation for its whole
lifetime. However, if the machine makes a mistake that we want to
correct, it is necessary for it to change its behavior--thus,
learning is a mandatory component.


Together with learning comes \emph{motivation}. Learning allows the
machine to adapt itself to the external environment, helping it to
produce outputs that maximize the function defined by its
motivation. Since we want to develop machines that make themselves
useful to humans, the motivation component should be directly
controlled by users through the communication channel. By specifying
positive and negative rewards, one may shape the behavior of the
machine so that it can become useful for concrete tasks (this is very
much in the spirit of reinforcement learning, \citealp[see,
e.g.,][]{Sutton:Barto:1998}, and discussion in Section
\ref{sec:related} below). 


Note that we will often refer to human learning as a source of insight
and an ideal benchmark to strive for. This is natural, since we would
like our machines to develop human-like intelligence. At the same
time, children obviously grow in a very different environment from the
one in which we tutor our machines, they soon develop a sophisticated
sensorimotor system to interact with the world, and they are innately
endowed with many other cognitive capabilities. An intelligent
machine, on the other hand, has no senses, and it will start its life
as a \emph{tabula rasa}, so that it will have to catch up not only on
human ontogeny, but also on their phylogeny (the history of AI
indicates that letting a machine learn from data is a more effective
strategy than manually pre-encoding ``innate'' knowledge into it). On
the positive side, the machine is not subject to the same biological
constraints of children, and we can, for example, expose it to
explicit tutoring at a rate that would not be tolerable for
children. Thus, while human learning can provide useful inspiration,
we are by no means trying to let our machines develop in human-like
ways, and we claim no psychological plausibility for the methods
we propose.

\section{A simulated ecosystem to educate com\-mu\-ni\-cat\-ion-based
  intelligent machines}
\label{sec:thegame}

In this section, we describe a simulated environment designed to teach
the basics of linguistic interaction to an intelligent machine, and how to use it to learn to
operate in the world. The simulated ecosystem should be seen as a
``kindergarten'' providing basic education to intelligent
machines. The machines are trained in this controlled environment to later
be connected to the real world in order to learn how to help humans with
their various needs.

The ecosystem I/O channels are controlled by an automatic mechanism,
avoiding the complications that would arise from letting the machine
interact with the ``real world'' from the very beginning, and allowing
us to focus on challenges that should directly probe the effectiveness
of new machine learning techniques.

The environment must be challenging enough to force the
machine to develop sophisticated learning strategies (essentially, it
should need to ``learn how to learn'').  At the same time, complexity
should be manageable, i.e., a human put into a similar environment
should not find it unreasonably difficult to learn to communicate and
act within it, even if the communication takes place in a language
the human is not yet familiar with. After mastering the basic
language and concepts of the simulated environment, the machine should
be able to interact with and learn from human teachers. %
This puts several restrictions on the kind of learning the machine
must come to be able to perform: most importantly, it will need to be
capable to extract the correct generalizations from just a few
examples, at a rate comparable to human learners.

Our ecosystem idea goes against received wisdom from the last decades
of AI research. This received wisdom suggests that systems should be
immediately exposed to real-world problems, so that they don't get
stuck into artificial ``blocks worlds'' \citep{Winograd:1971}, whose
experimenter-designed properties might differ markedly from those
characterizing realistic setups. Our strategy is based on the
observation, that we will discuss in Section
\ref{sec:computational-systems}, that current machine learning
techniques cannot handle the sort of genuinely incremental learning of
algorithms that is necessary for the development of intelligent
machines, because they lack the ability to store learned skills in
long-term memory and compose them. To bring about an advance in such
techniques, we have of course many choices. It seems sensible to pick
the simplest one. The environment we propose is sufficient to
demonstrate the deficiencies of current techniques, yet it is simple
enough that we can fully control the structure and nature of the tasks
we propose to the machines, make sure they have a solution, and use
them to encourage the development of novel techniques. Suppose we were
instead to work in a more natural environment from the very beginning,
for example from video input. This would impose large infrastructure
requirements on the developers, it would make data pre-processing a
big challenge in itself, and training even the simplest models would
be very time-consuming. Moreover, it would be much more difficult to
formulate interrelated tasks in a controlled way, and define the
success criterion. Once we have used our ecosystem to develop a system
capable of learning compositional skills from extremely sparse reward,
it should be simple to plug in more natural signals, e.g., through
communication with real humans and Internet access, so that the system
would learn how to accomplish the tasks that people really want it to
perform. 

The fundamental difference between our approach and classic AI
blocks worlds is that we do not intend to use our ecosystem to script an
exhaustive set of functionalities, but to teach the machine the
fundamental ability to \emph{learn how to efficiently learn} by creatively
combining already acquired skills. Once such machine gets connected
with the real world, it should quickly learn to perform any new task its Teacher
will choose. Our environment can be seen as analogous to
explicit schooling. Pupils are taught math in primary school through
rather artificial problems. However, once they have interiorized basic
math skills in this setup, they can quickly adapt them to
the problems they encounter in their real life, and rely on them to rapidly
acquire more sophisticated mathematical techniques.

\subsection{High-level description of the ecosystem}
\label{sec:high-level-ecosystem}

\paragraph{Agents} To develop an artificial system that is able to
incrementally acquire new skills through linguistic interaction, we
should not look at the training data as a static set of labeled
examples, as in common machine learning setups. We propose instead a
dynamic ecosystem akin to that of a computer game. The Learner (the
system to be trained) is an actor in this ecosystem.

The second fundamental agent in the ecosystem is the Teacher. The
Teacher assigns tasks and rewards the Learner for desirable behaviour,
and it also provides helpful information, both spontaneously and in
response to Learner's requests. The Teacher's behaviour is entirely
scripted by the experimenters. Again, this might be worryingly
reminiscent of entirely hand-coded good-old AIs. However, the Teacher
need not be a very sophisticated program. In particular, for each task
it presents to the learner, it will store a small set of expected
responses, and only reward the Learner if its behaviour exactly
matches one response. Similarly, when responding to Learner's
requests, the Teacher is limited to a fixed list of expressions it
knows how to respond to. The reason why this suffices is that the aim
of our ecosystem is to kickstart the Learner's efficient learning
capabilities, and not to provide enough direct knowledge for it to be
self-sufficient in the world. For example, given the limitations of the
scripted Teacher, the Learner will only be able to acquire a very
impoverished version of natural language in the ecosystem. At the same
time, the Learner should acquire powerful learning and generalization
strategies. Using the minimal linguistic skills and strong learning
abilities it acquired, the Learner should then be able to extend its
knowledge of language fast, once it is put in touch with actual human
users.

Like in classic text-based adventure games \citep{Wikipedia:InteractiveFiction:2015}, the Environment is
entirely linguistically defined, and it is explored by the Learner by
giving orders, asking questions and receiving feedback (although
graphics does not play an active role in our simulation, it is
straightforward to visualize the 2D world in order to better track
the Learner's behaviour, as we show through some examples below). The
Environment is best seen as the third fundamental agent in the
ecosystem. The Environment behaviour is also scripted. However, since
interacting with the Environment serves the purpose of observation and
navigation of the Learner surroundings (``sensorimotor experience''),
the Environment uses a controlled language that, compared to that of
the Teacher, is more restricted, more explicit and less ambiguous. One
can thus think of the Learner as a higher-level programming
language, that accepts instructions from the programmer (the Teacher)
in a simple form of natural language, and converts them into the machine code
understood by the Environment.

In the examples to follow, we assume the world defined by the Environment to be split into discrete
cells that the Learner can traverse horizontally and vertically. The
world includes barriers, such as walls and water, and a number of
objects the Learner can interact with (a pear, a mug, etc).

Note that, while we do not explore this possibility here, it might be
useful to add other actors to the simulation: for example, training
multiple Learners in parallel, encouraging them to teach/communicate
with each other, while also interacting with the scripted Teacher.


\paragraph{Interface channels} The Learner experience is entirely
defined by generic \emph{input} and \emph{output} channels. The
Teacher, the Environment and any other language-endowed agent write to
the input stream. Reward (a scalar value, as discussed next) is also
written to the input stream (we assume, however, that the Learner does
not need to discover which bits encode reward, as it will need this
information to update its objective function). Ambiguities are avoided
by prefixing a unique string to the messages produced by each actor
(e.g., messages from the Teacher might be prefixed by the string
\textbf{T:}, as in our examples below). The Learner writes to its
output channel, and it is similarly taught to use unambiguous prefixes
to address the Teacher, the Environment and any other agent or service
it needs to communicate with. Having only generic input and output
communication channels should facilitate the seamless addition of new
interactive entities, as long as the Learner is able to learn the
language they communicate in.

\paragraph{Reward} Reward can be positive or negative (1/-1), the
latter to be used to speed up instruction by steering away the Learner
from dead ends, or even damaging behaviours.  The Teacher, and later
human users, control reward in order to train the Learner. We might
also let the Environment provide feedback through hard-coded rewards,
simulating natural events such as eating or getting hurt. Like in
realistic biological scenarios, reward is sparse, mostly being
a\-war\-ded after the Learner has accomplished some task.  As
intelligence grows, we expect the reward to become \emph{very} sparse,
with the Learner able to elaborate complex plans that are only
rewarded on successful completion, and even displaying some degree of
self-motivation.  Indeed, the Learner should be taught that short-term
positive reward might lead to loss at a later stage (e.g., hoarding on
food with poor nutrition value instead of seeking further away for
better food), and that sometimes reward can be maximized by engaging
in activities that in the short term provide no benefit (learning to
read might be boring and time-consuming, but it can enormously speed
up problem solving--and the consequent reward accrual-- by making the
Learner autonomous in seeking useful information on the
Internet). Going even further, during the Learner ``adulthood''
explicit external reward could stop completely. The Learner will no
longer be directly motivated to learn in new ways, but ideally the
policies it has already acquired will include strategies such as
curiosity (see below) that would lead it to continue to acquire new
skills for its own sake.  Note that, when we say that reward could
stop completely, we mean that users do not need to provide explicit
reward, in the form of a scalar value, to the Learner. However, from a
human perspective, we can look at this as the stage in which the
Learner has interiorized its own sources of reward, and no longer
needs external stimuli.

We assume binary reward so that human users need not worry about relative \emph{amounts} of reward to give to the Learner %
(if they do want to control the amount of reward, they can
simply reward the Learner multiple times). The Learner objective
should however maximize \emph{average reward over time}, naturally
leading to different degrees of cumulative reward for different
cour\-ses of action (this is analogous to the notion of expected
cumulative reward in reinforcement learning, which is a possible way
to formalize the concept). Even if two solutions to a task are
rewarded equally on its completion, the faster strategy will be
favored, as it leaves the Learner more time to accumulate further
reward. This automatically ensures that efficient solutions are
preferred over wasteful ones. Moreover, by measuring time
independently from the number of simulation steps, e.g., using simple
wall-clock time, one should penalize inefficient learners spending a
long time performing offline computations.

As already mentioned, our approach to reward-based learning shares
many properties with reinforcement learning. Indeed, our setup fits
into the general formulation of the reinforcement learning problem
\citep{kaelbling:etal:1996,Sutton:Barto:1998}--see Section
\ref{sec:related} for further discussion of this point.

\paragraph{Incremental structure} In keeping with the game idea, it is
useful to think of the Learner as progressing through a series of
levels, where skills from earlier levels are required to succeed in
later ones. Within a level, there is no need to impose a strict
ordering of tasks (even when our intuition suggests a natural
incremental progression across them), and we might let
the Learner discover its own optimal learning path
by cycling multiple times through blocks of them.

At the beginning, the Teacher trains the Learner to perform very
simple tasks in order to kick-start linguistic communication and the
discovery of very simple algorithms. The Teacher first rewards the
Learner when the latter repeats single characters, then words,
delimiters and other control strings. The Learner is moreover taught
how to repeat and manipulate longer sequences. In a subsequent block of tasks,
the Teacher leads the Learner to develop a semantics for linguistic
symbols, by encouraging it to associate linguistic expressions with
actions. This is achieved through practice sessions in which the
Learner is trained to repeat strings that function as Environment
commands, and it is rewarded only when it takes notice of the effect
the commands have on its state (we present concrete examples
below). At this stage, the Learner should become able to associate
linguistic strings to primitive moves and actions (\emph{turn left}). Next, the Teacher will assign tasks involving action sequences
(\emph{find an apple}), and the Learner should convert them into
sets of primitive commands (simple ``programs''). The Teacher will,
increasingly, limit itself to specify an abstract end goal
(\emph{bring back food}), but not recipes to
accomplish it, in order to spur creative thinking on behalf of the
Learner (e.g., if the Learner gets trapped somewhere while looking for
food, it may develop a strategy to go around obstacles).
In the process of learning to parse and
execute higher-level commands, the Learner should also be trained to
ask clarification questions to the Teacher (e.g., by initially
granting reward when it spontaneously addresses the Teacher, and by
the repetition-based strategy we illustrate in the examples
below). With the orders becoming more general and complex, the
language of the Teacher will also become (within the limits of what
can be reasonably scripted) richer and more ambiguous, challenging the
Learner capability to handle restricted specimens of common natural
language phenomena such as polysemy, vagueness, anaphora and
quantification. %

To support user scenarios such as the ones we envisaged in Section
\ref{sec:desiderata} above and those we will discuss at the end of
this section, the Teacher should eventually teach the Learner how to
``read'' natural text, so that the Learner, given access to the
Internet, can autonomously seek for information online. Incidentally,
notice that once the machine can read text, it can also exploit distributional
learning from large amounts of text \citep{Erk:2012,Mikolov:etal:2013b,Turney:Pantel:2010}
to induce word and phrase representations addressing some of the
challenging natural language phenomena we just mentioned, such as
polysemy and vagueness.

The Learner must take its baby
steps first, in which it is carefully trained to accomplish simple tasks
such as learning to compose basic commands. However, for the Learner
to have any hope to develop into a fully-functional intelligent
machine, we need to aim for a ``snow-balling'' effect to soon take place, such
that later tasks, despite being inherently more complex, will require
a lot less explicit coaching, thanks to a combinatorial explosion in
the background abilities the Leaner can creatively compose (like for
humans, learning how to surf the Web should take less time than
learning how to spell).

\paragraph{Time off} Throughout the simulation, we
foresee phases in which the Learner is free to interact with the
Environment and the Teacher without a defined task. Systems should
learn to exploit this time off for undirected exploration, that should
in turn lead to better performance in active training stages, just like,
in the dead phases of a video-game, a player is more likely to try out
her options than to just sit waiting for something to happen, or when
arriving in a new city we'd rather go sightseeing than staying in the
hotel. Since curiosity is beneficial in many situations, such
behaviour should naturally lead to higher later rewards, and thus be
learnable. Time off can also be used to ``think'' or ``take a nap'', in
which the Learner can replay recent experiences and possibly update its
inner structure based on a more global view of the knowledge it has accumulated,
given the extra computational resources that the free time policy offers.




\paragraph{Evaluation} Learners can be quantitatively evaluated and
compared in terms of the number of new tasks they accomplish successfully in a fixed amount of time, 
a measure in line with the reward-maximization-over-time objective we
are proposing. %
Since the interactive, multi-task environment setup does not
naturally support a distinction between a training and a test phase,
the machine must carefully choose reward-maximizing actions from the
very beginning. 
In contrast, evaluating the machine only on its final behavior would
overlook the number of attempts it took to reach the solution.  Such
alternative evaluation would favor models which are simply able to
memorize patterns observed in large amounts of training data. In many
practical domains, this approach is fine, but we are interested in
machines capable of learning truly general problem-solving
strategies. As the tasks become incrementally more difficult, the
amount of required computational resources for naive
memorization-based approaches scales exponentially, so only a machine
that can efficiently generalize can succeed in our environment. We
will discuss the limitations of machines that rely on memorization
instead of algorithmic learning further in Section
\ref{sec:computational} below.

We would like to foster the development of intelligent machines by
employing our ecosystem in a public competition. Given what we just
said, the competition would not involve distributing a static set of
train\-ing/develop\-ment data similar in nature to the final test
set. We foresee instead a setup in which developers have access to the
full pre-programmed environment for a fixed amount of time. The
Learners are then evaluated on a set of new tasks that are
considerably different from the ones exposed in the development phase.
Examples of how test tasks might differ from those encountered during
development include the Teacher speaking a new language, a different
Environment topography, new obstacles and objects with new
affordances, and novel domains of endeavor (e.g., test tasks might
require selling and buying things, when the Learner was not previously
introduced to the rules of commerce). 

\subsection{Early stages of the simulation}

\paragraph{Preliminaries} At the very beginning, the Learner has to
learn to pay attention to the Teacher, to identify the basic units of
language (find regularity in bit patterns, learn characters, then
words and so on). It must moreover acquire basic sequence repetition and
manipulation skills, and develop skills to form memory and learn efficiently.
These very initial stages of learning are extremely important, as we believe they
constitute the building blocks of intelligence.

However, as bit sequences do not make for easy readability, we focus here on an
immediately following phase, in which the Learner has already learned
how to pay attention to the Teacher and manipulate character
strings. We show how the Teacher guides the Learner from these basic skills to
being able to solve relatively sophisticated Environment navigation
problems by exploiting interactive communication. Because of the
``fractal-like'' structure we envisage in the acquisition of
increasingly higher-level skills, these steps will illustrate many of
the same points we could have demonstrated through the lower-level
initial routines. The tasks we describe are also incrementally
structured, starting with the Learner learning to issue Environment
commands, then being led to take notice of the effect these commands
have, then understanding command structure, in order to generalize
across categories of actions and objects, leading it in turn to
being able to process higher-level orders. At this point, the Learner
is initiated to interactive communication.

Note that we only illustrate here ``polite'' turn-taking, in which
messages do not overlap, and agents start writing to the communication
channels only after the end-of-message symbol has been issued. We do
not however assume that interaction must be constrained in this
way. On the contrary, there are advantages in letting entities write
to the communication channels whenever they want: for example, the
Teacher might interrupt the Learner to prevent him from completing a
command that would have disastrous consequences, or the Learner may
interrupt the Teacher as soon as it figured out what to do, in order
to speed up reward (a simple priority list can be defined to solve
conflicts, e.g., Teacher's voice is ``louder'' than that of
Environment, etc.).

Note also that our examples are meant to illustrate specific instances
from a larger set of trials following similar templates, that should
involve a variety of objects, obstacles and possible
actions. Moreover, the presented examples do not aim to exhaustively
cover all learning-fostering strategies that might be implemented in
the ecosystem. Finally, we stress again that we are not thinking of a
strict ordering of tasks (not least because it would be difficult to
fix, \emph{a priori}, an ordering that is based on some objective,
learning-model-independent difficulty criterion), but rather about
clusters of related tasks organized into levels, such that the Learner
could cycle through them and discover its own optimal progression to
solve them.

\paragraph{Notation} We present the input to the
Learner (messages from Teacher and Environment, as well as Reward,
prefixed by \textbf{T:}, \textbf{E:} and \textbf{R:}, respectively) in
the left column, and the Learner output on the right (messages
addressed to the Teacher are prefixed by \textbf{@T:}, those for the
Environment are prefixed by \textbf{@E:}). We assume different I/O
identification prefixes to disambiguate meta-linguistic statements
(e.g., ``\textbf{T:} give order \textbf{@E:} I move.'' in the
first exchange below, where it's only the Teacher speaking all the
way). The full stop serves as end-of-message delimiter (line breaks
are used for formatting purposes, and should be treated as white
space; similarly, hyphens are inserted for readability and should be
ignored).  We use an ellipsis (three dots) to mark a 
sequence of messages on any channel we do not report (e.g., the Learner
exploring a number of solutions before zeroing in on the right
one). Finally, we interleave the conversations with comments, shown in
grey.

\paragraph{The Learner learns to issue Environment commands} We start
with the Teacher teaching the Learner to produce Environment commands,
first by rote repetition channeled to the right message recipient
(\textbf{@E}), then by encouraging it to translate from the Teacher's
own natural language to the controlled Environment language. %
At this stage, reward is not tied to understanding what happened (the
``semantics'' of the commands). Still, a good Learner might take note
of the actions triggered by the commands. The following are example
conversations that may happen during the learning process:

% ...
\begin{longtable}{p{0.45\linewidth}p{0.45\linewidth}}
  \emph{Input:}&  \emph{Output:}\\
  \endhead
  \textbf{T:} give order \textbf{@E:} I move.&\\
  &fdakjf13ktj53u9gs5 5 3\#.\\
  \CLCOMMENT{character soup without addressee, no reward}
  \multicolumn{2}{c}{\ldots}\\
  &\textbf{@T:} dsfjkasjfk;aljsf.\\
  \CLCOMMENT{character soup with wrong addressee, no reward}
  \CELAPSE{}
  &\textbf{@E:} fglk4\$3wfgg.\\
  \CLCOMMENT{right addressee, character soup, no reward} \CELAPSE{}
  &give order \textbf{@E:} I move.\\
  \textbf{E:} you moved.&\\
  \textbf{R:} 1.&\\
  \CLCOMMENT{Learner moves and is rewarded, although the solution is inefficient}
  \textbf{T:} give order \textbf{@E:} I move.&\\
  \CELAPSE{}
  &\textbf{@E:} I move.\\
  \textbf{E:} you moved.&\\
  \textbf{R:} 1.&\\
  \CLCOMMENT{Learner figured out minimal string leading to reward}
\end{longtable}

Further skills the Learner will need to develop include segmentation
of Teacher's input, and production of separate commands for the
Environment.   At first, this may be rather trivial, as shown in this
example:

\begin{longtable}{p{0.45\linewidth}p{0.45\linewidth}}
  \emph{Input:}&  \emph{Output:}\\
  \endhead
  \textbf{T:} give orders \textbf{@E:} I move and \textbf{@E:} I look.&\\
  &\textbf{@E:} I move and \textbf{@E:} I look.\\
  \CLCOMMENT{no reward, because Environment does not parse complex commands} \CELAPSE
  \textbf{T:} give orders \textbf{@E:} I move and \textbf{@E:} I look.&\\
    &\textbf{@E:} I move.\\
    \textbf{E:} you moved.&\\
    &\textbf{@E:} I look.\\
    \textbf{E:} there is an apple.&\\
    \textbf{R:} 1.&\\
\end{longtable}

\begin{figure}[p]
\begin{center}
\begin{tabular}{lr|lr}
\multicolumn{2}{c}{\includegraphics[width=.3\linewidth]{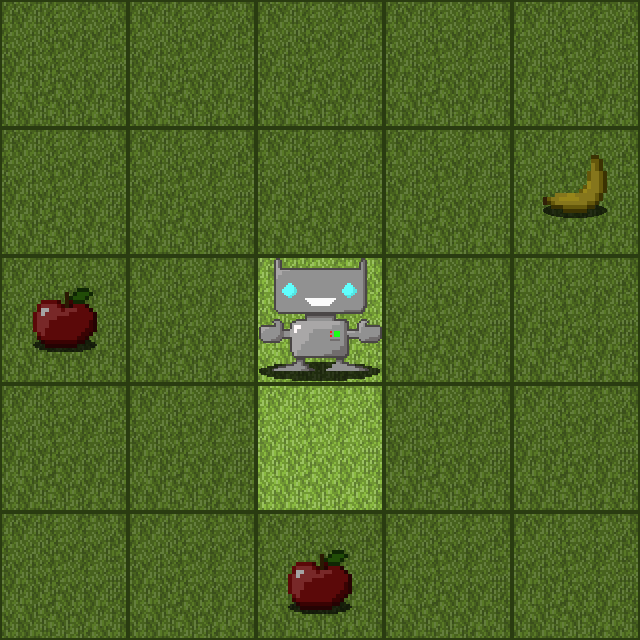}}
&
\multicolumn{2}{c}{\includegraphics[width=.3\linewidth]{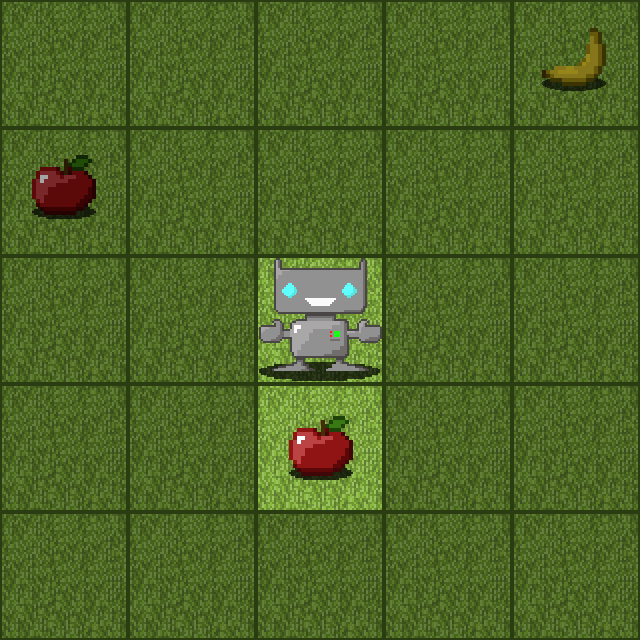}}
\\
\emph{Input:}& \emph{Output:} & \emph{Input:}& \emph{Output:}
\\
\textbf{T:} move and look. & & \textbf{E}: you moved.& \\
  &\textbf{@E:} I move. & &\textbf{@E:} I look.
\\
& & \textbf{E:} there is an apple.&\\
& & \textbf{R:} 1.&\\
\end{tabular}
\end{center}
\caption{Example of how the simulation might be visualized to help
  developers track Learner's behaviour. The left panel represents the
  Environment at the stage in which Learner issues the move command,
  the right panel depicts the Environment after this command is
  executed. A lighter shade of green marks the cell the Learner
  occupies, and the one it is turned towards. These cells are directly observable.
  (Best viewed in color.)}\label{fig:move-and-look}
\end{figure}

Eventually, the Learner should learn to switch fluently
between Teacher and Environment communication codes, translating the
Teacher's linguistic messages into motor commands to the Environment
(an example of such exchange is visualized in Figure
\ref{fig:move-and-look}).

\begin{figure}[p]
\makebox[\textwidth]
{\footnotesize
\begin{tabular}{lr|lr|lr}
\multicolumn{2}{c}{\includegraphics[width=.3\linewidth]{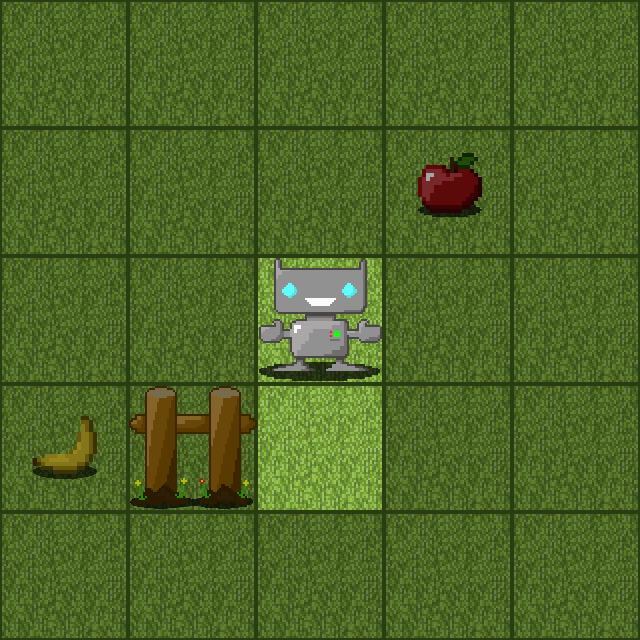}}
&
\multicolumn{2}{c}{\includegraphics[width=.3\linewidth]{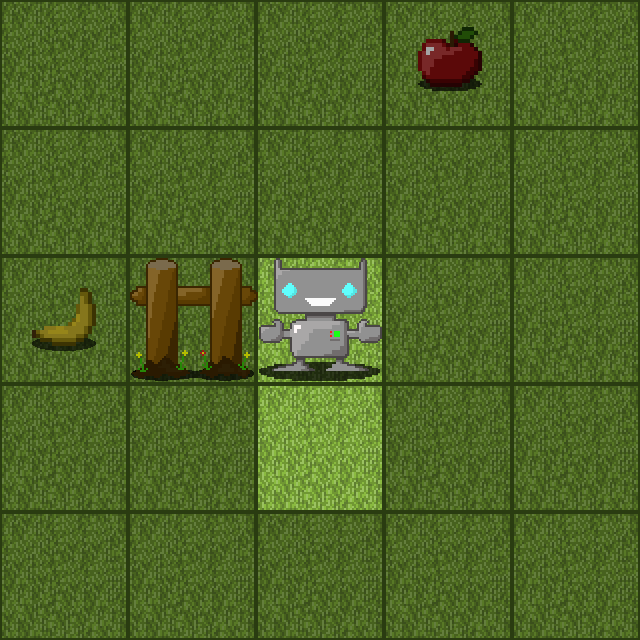}}
&
\multicolumn{2}{c}{\includegraphics[width=.3\linewidth]{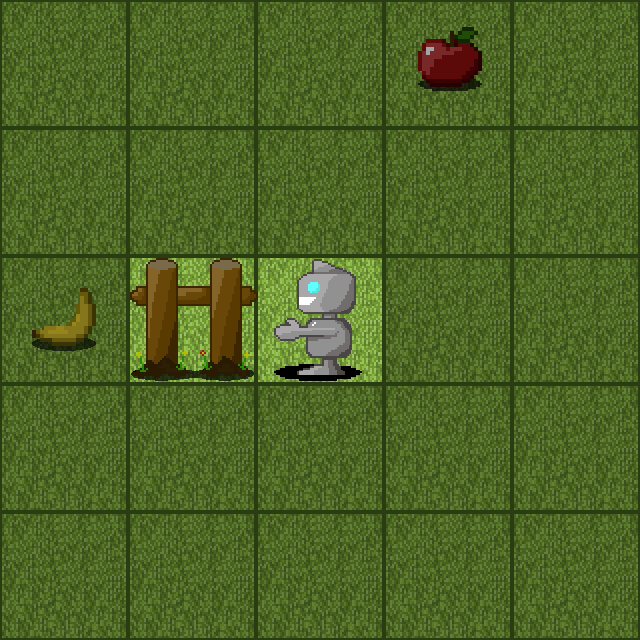}}
\\
\emph{Input:}     & \emph{Output:}                   & \emph{Input:}        & \emph{Output:}               &\emph{Input:}              & \emph{Output:}
\\
\multicolumn{2}{l|}{\textbf{T:} move, turn right and move.} & \multicolumn{2}{l|}{\textbf{E:} you moved.}          &\multicolumn{2}{l}{\textbf{E}: you turned right.}
\\
\multicolumn{2}{r|}{\textbf{@E:} I move.}             & \multicolumn{2}{r|}{\textbf{@E:} I turn right.}      &\multicolumn{2}{r}{\textbf{@E:} I move.}
\\
                 &                     &                     &                             & \textbf{E:} you can't move.&
\\
\end{tabular}
}
\caption{The Teacher gives the order \emph{move, turn right and move}. The
  Learner is however blocked by a barrier after it
  turns right, and reward is delayed, because the order is not executed.  
  (Best viewed in color.)}\label{fig:move-and-stop}
\end{figure}





\paragraph{Associating language to actions} Next, the Learner is
encouraged to take notice of what happens in the Environment and
associate the Teacher's language to states of the world, by being
rewarded only when its orders result in the corresponding actions
being effectively carried out in the world. An example task is
illustrated in Figure \ref{fig:move-and-stop}, where the Learner will
not get reward because it is no longer sufficient for it to correctly translate
instructions into Environment commands: the commands must also have
the intended effect on the world. In this specific case, order
execution fails because the Learner encountered a barrier.



\paragraph{Learning to generalize} By adding sufficient variety to the
trials above, the Learner should start to get hold of the
compositional nature of commands (\emph{turn right} and \emph{turn
  left} share systematic characteristics, e.g., they both cause a
change in what the Learner sees; the Learner can pick X only when X is
in the cell in front of him, etc.). The next trials help the Learner further move
away from literal command string memorization, and generalize across
categories of actions and objects (\emph{turn left/right} are
instances of \emph{turning}, \emph{apples} and \emph{pears} are
\emph{objects}, etc.). This is demonstrated in the following example:

\begin{longtable}{p{0.45\linewidth}p{0.45\linewidth}}
  \emph{Input:}&  \emph{Output:}\\
  \endhead
\textbf{T:} turn and move.&\\
&\textbf{@E:} I turn.\\
\textbf{E:} \it{(no action)}&\\
\CELAPSE
&\textbf{@E:} I turn left.\\
\textbf{E:} you turned left.&\\
&\textbf{@E:} I move.\\
\textbf{E:} you can't move.&\\
\CLCOMMENT{there is a barrier in front of Learner}
\CELAPSE
&\textbf{@E:} I turn right.\\
\textbf{E:} you turned right.&\\
&\textbf{@E:} I move.\\
\textbf{E:} you moved.&\\
\textbf{R:} 1.&\\
\end{longtable}

In the next example, the Learner is asked to pick some object that is
in front of him, without specifying what the object is (we assume that
the simple, unambiguous language of the Environment has no words for
general categories such as that of object, so the Learner must
eventually find out that it must name specific objects instead):

\begin{longtable}{p{0.45\linewidth}p{0.45\linewidth}}
  \emph{Input:}&  \emph{Output:}\\
  \endhead
\textbf{T:} pick an object.&\\
&\textbf{@E:} I pick an object.\\
\textbf{E:} \it{(no response)}&\\
&\textbf{@E:} I pick the apple.\\
\textbf{E:} \it{(no response)}&\\
&\textbf{@E:} I pick the pear.\\
\textbf{E:} you picked the pear.&\\
\textbf{R:} 1.&\\
\end{longtable}

As we just illustrated, initially the Learner will apply an exhaustive
search strategy, listing all objects it knows of to find one it can
pick.  Later, the Teacher should teach the \emph{look} command, and
the Learner should be able to discover a faster strategy than
exhaustive search:

\begin{longtable}{p{0.45\linewidth}p{0.45\linewidth}}
  \emph{Input:}&  \emph{Output:}\\
  \endhead
\textbf{T:} pick an object.&\\
&\textbf{@E:} I look.\\
\textbf{E:} you see a pear.&\\
&\textbf{@E:} I pick the pear.\\
\textbf{E:} you picked the pear.&\\
\textbf{R:} 1.&\\
\end{longtable}

Learning new strategies of course does not imply that the Learner can safely
forget the ones it has previously acquired. For example, in some tasks
the \emph{look} command might not work (because, say, it is too dark
and the Learner cannot see what is in front of him). In such case, an
efficient Learner should find it easy to re-use previously learned
skills, such as exhaustive search.

\paragraph{Understanding higher-level orders} The next batch of trials
aims at developing the ability to decompose higher-level commands into
a sequence of basic actions, first fixed (\emph{move twice}), then
flexible (\emph{find an apple}). The general teaching strategy is to
provide sets of paired trials: In the first trial, the Teacher
describes the task explicitly as a sequence of primitive actions, and
gives the high-level name of the order. In the second trial, the Teacher
issues an equivalent high-level order, and the Learner is rewarded if
it goes through the same sequence as in the first trial (we are aware
that many complex linguistic expressions are not amenable to a
straightforward decompositional analysis, see, e.g.,
\citealp{Boleda:Erk:2015}: we will use some that are in order to teach
the Learner about how to decompose complex concepts into step-by-step
plans). The Teacher could start by teaching iteration counters:

\begin{longtable}{p{0.45\linewidth}p{0.45\linewidth}}
  \emph{Input:}&  \emph{Output:}\\
  \endhead
\textbf{T:} move and move.&\\
&\textbf{@E:} I move.\\
\textbf{E:} you moved.&\\
&\textbf{@E:} I move.\\
\textbf{E:} you moved.&\\
\textbf{R:} 1.&\\
\textbf{T:} this is called move two times.&\\
\CELAPSE
\textbf{T:} move two times.&\\
\CELAPSE
&\textbf{@E:} I move.\\
\textbf{E:} you moved.&\\
&\textbf{@E:} I move.\\
\textbf{E:} you moved.&\\
\textbf{R:} 1.&\\
\CELAPSE
\textbf{T:} move and move and move.&\\
&\textbf{@E:} I move.\\
\textbf{E:} you moved.&\\
&\textbf{@E:} I move.\\
\textbf{E:} you moved.&\\
&\textbf{@E:} I move.\\
\textbf{E:} you moved.&\\
\textbf{R:} 1.&\\
\textbf{T:} this is called move three times.&\\
\CELAPSE
\textbf{T:} move three times.&\\
\CELAPSE
&\textbf{@E:} I move.\\
\textbf{E:} you moved.&\\
&\textbf{@E:} I move.\\
\textbf{E:} you moved.&\\
&\textbf{@E:} I move.\\
\textbf{E:} you moved.&\\
\textbf{R:} 1.&\\
\end{longtable}

The Learner should eventually be able to apply modifiers productively
to different actions without much further training (e.g., apply
\emph{three times} to \emph{turn left}).


Next, the Teacher can  show how to segment high-level
tasks, such as \emph{finding something}, into atomic action
sequences. For example, to find an apple, one can develop a simple
strategy of going forward until an apple is found. 
The Teacher might initiate the Learner to this new skill as follows:

\begin{longtable}{p{0.45\linewidth}p{0.45\linewidth}}
  \emph{Input:}&  \emph{Output:}\\
  \endhead
\textbf{T:} move and look.&\\
&\textbf{@E:} I move.\\
\textbf{E:} you moved.&\\
&\textbf{@E:} I look.\\
\textbf{E:} you see grass.&\\
\textbf{T:} move and look.&\\
&\textbf{@E:} I move.\\
\textbf{E:} you moved.&\\
&\textbf{@E:} I look.\\
\textbf{E:} you see an apple.&\\
\textbf{R:} 1.&\\
\textbf{T:} this is called find an apple.&\\
\end{longtable}

Note that one such example is not sufficient for the Learner to
understand what exactly we call \emph{finding an apple}, as multiple
interpretations are valid: maybe we just wanted it to execute the
given commands twice. Thus, there should be multiple training
sequences of variable length, to clarify that the task is really to
apply an \texttt{until} loop, i.e., to iterate \emph{move} and
\emph{look} commands until the object the Learner is searching for is
found.

Further tasks can define composite skills, such as \emph {getting an
  apple}, which would consist of first finding an apple, and then
picking it up. Another generalization would be to include multiple
objects the Learner should be searching for.  Adding obstacles to the
Environment would further complicate the challenge.  The previous
strategy of simply going forward until an object is found will not
work anymore, as the Learner would stop at the first obstacle. We can
either expect the Learner to independently develop better search
strategies involving turning, or we can add further examples where the
Teacher shows the Learner how to improve through more direct
supervision.

\paragraph{Interactive communication} Tasks such as finding an
  apple without directions might involve a long random search. Thus, we want to
kick-start interactive communication, so that the Learner can be
efficiently directed by the Teacher (and eventually by humans) to the
right cell. In a first set of trials, the Learner is rewarded for
repeating a \emph{how to} request uttered by the Teacher (addressing
it back to the teacher via the \textbf{@T:} prefix), and following the
precise instructions produced by the Teacher in response to the
request:

\begin{longtable}{p{0.45\linewidth}p{0.45\linewidth}}
  \emph{Input:}&  \emph{Output:}\\
  \endhead
\textbf{T:} ask me how to find an apple.&\\
&\textbf{@E:} ask me how to find an apple.\\
\CLCOMMENT{wrong addressee, wrong request prefix}
\CELAPSE
&\textbf{@T:} how to find an apple.\\
\textbf{T:} turn right and move and move and pick the apple.&\\
&\textbf{@E:} I turn right.\\
\textbf{E:} you turned right.&\\
&\textbf{@E:} I move.\\
\textbf{E:} you moved.&\\
&\textbf{@E:} I move.\\
\textbf{E:} you moved.&\\
&\textbf{@E:} I pick the apple.\\
\textbf{E:} you picked the apple.&\\
\textbf{R:} 1.&\\
\end{longtable}

Trials such as this one are later interspersed with trials where the
Learner is assigned a task it can in principle accomplish by random
search, but taking the initiative by issuing a \emph{how to} request
and then following the precise directions provided by the Teacher will
considerably speed up reward.


\paragraph{Algorithmic knowledge} Some tasks illustrated above require
understanding basic control flow structures. For example, parsing action
modifiers implies a simple form of counting, and in order to find
things the Learner must implement an \texttt{until} (equivalently,
\texttt{while not}) loop. Similarly, the command \emph{get out of the
  grass} calls for a \texttt{while} loop. Efficient completion of more
advanced tasks, e.g., \emph{return home},
implies development of more complex algorithms, such as
path-finding. After acquiring a bank of such algorithms (which might be encoded in the internal machine representation by something very different from programming language constructs), the Learner
should be able, in advanced stages of the simulation, to productively
combine them in order to succeed in full-fledged novel missions that
involve accomplishing a large number of hierarchically-structured
sub-goals (\emph{find somebody who will trade two apples for a banana}). 

As we discussed in Section \ref{sec:high-level-ecosystem}, the Learner's
functionality could essentially be interpreted as learning how to compose programs based on the
descriptions given in natural language by the Teacher. The programs
produce very simple instructions that are understood by the
Environment, which can be seen as a sort of CPU. From this point of
view, the intelligent system we aim to train is a bridge between the
Teacher (later to be replaced by a human operator) and a traditional
computer that understands only a limited set of basic commands and
needs to be manually programmed for each single task. Thus, we believe
that successful construction of intelligent machines could automate
computer programming, which will likely be done in the future simply
through communication in natural language.

\subsection{Interacting with the trained intelligent machine}
\label{sec:advanced-examples}

To conclude the illustration of our plan, we provide a motivating example of how an intelligent machine
schooled in our ecosystem could later make itself useful in the real
world. We consider a scenario in which the machine works as an
assistant to Alice, an elderly person living alone. Bob is Alice's
son, and he also interacts with the machine.

We assume that, as part of its training, the machine has been taught
how to issue Internet commands and process their outcomes. 
In the example dialogue, we give a general idea
of how the machine would interface to the Internet, without attempting
to precisely define the syntax of this interaction. Most importantly,
the Internet queries in the example are meant to illustrate how the
machine does not need to store all the knowledge it needs to
accomplish its duties, as it can retrieve useful information from the
Web on demand, and reason about it.

\begin{longtable}{p{0.45\linewidth}p{0.45\linewidth}}
  \emph{Input:}&  \emph{Output:}\\
  \endhead
  \textbf{Bob:} I just spoke to the doctor, who said my mother needs to move for at least one hour per day, please make sure she does get enough exercise.&\\
  \CELAPSE
  \CLCOMMENT{following conversation takes place the next day:}
  &\textbf{@Alice:} Are you in the mood for some light physical exercise today?\\
  \textbf{Alice:} Yes, but no more than 15 minutes, please.&\\
  &\textbf{@INTERNET:} [query search engine for keywords \emph{elderly, light activity, 15 minutes}]\\
  \CLCOMMENT{shortly afterwards\ldots}
  &\textbf{@Alice:} I have downloaded a YouTube video with a 15-minute yoga routine, please watch it whenever you can.\\   
  \CLCOMMENT{a few hours later\ldots}
  \textbf{Alice:} I'm going out to buy groceries.&\\
  &\textbf{@INTERNET:} [query search engine with keywords \emph{average walking speed, elderly person}]\\
  &\textbf{@INTERNET:} [search maps app for distance to grocery stores in Alice's neighborhood]\\
  &\textbf{@Alice:} Hey, why don't you walk to the Yummy Food Market today? It should take you about 45 minutes to and from, so you'll get the rest of your daily exercise.\\
  \textbf{@Alice:} Thanks for the suggestion.&\\
\end{longtable}


The communication-based intelligent machine should adapt to a
whole range of tasks it was not explicitly programmed for.
If necessary, the user can give it further
explicit positive and negative reward to motivate it to change its
behavior. This may be needed only rarely, as the machine should be
shipped to the end users after it already mastered good communication
abilities, and further development should mostly occur through
language. For example, when the user says \emph{No, don't do this
  again}, the machine will understand that repeating the same type of
behavior might lead to negative reward, and it will change its course
of action even when no explicit reward signal is given (again, another way to put this is that the machine should associate similar linguistic strings to an ``interiorized'' negative reward).

The range of tasks for intelligent machines can be very diverse:
besides the everyday-life assistant we just considered, it could
explain students how to accomplish homework assignments, gather
statistical information from the Internet to help medical researchers (see also the examples in Section \ref{sec:communication} above), find bugs in computer programs, or even write programs on its own. Intelligent machines should extend our intellectual abilities in the same way current computers already function as an extension to our memory. This should enable us to perform intellectual tasks beyond what is possible today.

We realize the intelligent machines we aim to construct could become powerful tools that
may be possibly used for dubious purposes (the same could be said about any advanced technology,
including airplanes, space rockets and computers). We believe the perception of AI is skewed by popular
science fiction movies. Instead of thinking of computers that take over the world for their own
reasons, we think AI will be realized as a tool: A machine that will extend our capability to reason and solve
complex problems. Further, given the current state of the technology, we believe any discussion on
``friendliness'' of the AI is at this moment premature. We expect it will take years, if not decades
to scale basic intelligent machines to become competitive with humans, giving us enough time to discuss
any possible existential threats.

\section{Towards the development of intelligent machines}
\label{sec:computational-systems}

In this section, we will outline some of our ideas about
how to build intelligent machines that would benefit from the learning
environment we described. While we do not have a concrete proposal yet
about how exactly such machines should be implemented, we will discuss
some of the properties and components we think are needed to support
the desired functionalities. We have no pretense of completeness, we
simply want to provide some food for thought. As in the previous
sections, we try to keep the complexity of the machine at the minimum,
and only consider the properties that seem essential.

\subsection{Types of learning}

There are many types of behavior that we collectively call learning,
and it is useful to discuss some of them first.
Suppose our goal is to build an intelligent machine
working as a translator between two languages (we take here a simplified word-based view of the translation task).  First, we will teach
the machine basic communication skills in our simulated environment so
that it can react to requests given by the user.  Then, we will start
teaching it, by example, how various words are translated.

There are different kinds of learning happening here. To master 
basic communication skills, the machine will have to understand the
concept of positive and negative reward, and develop complex
strategies to deal with novel linguistic inputs. This requires
discovery of algorithms,
 and the ability
to remember facts, skills and even learning strategies.

Next, in order to translate, the machine needs to store pairs of
words. The number of pairs is unknown and a flexible growing mechanism
may be required. 
However,
once the machine understands how to populate the dictionary with
examples, the learning left to do is of a very simple nature: the
machine does not have to update its learning strategy, but only to
store and organize the incoming information into long-term memory
using previously acquired skills. Finally, once the vocabulary
memorization process is finished and the machine starts working as a
translator, no further learning might be required, and the functionality of the machine can be fixed.

The more specialized and narrow the functionality of the machine is,
the less learning is required. For
very specialized forms of behavior, it should be possible to program the
solution manually. However, as we move from roles such as a simple
translator of words, a calculator, a chess player, etc., to machines
with open-ended goals, we need to rely more on general learning from a
limited number of examples.

One can see the current state of the art in machine learning as being
somewhere in the middle of this hierarchy. Tasks such as automatic
speech recognition, classification of objects in images or machine
translation are already too hard to be solved purely through manual
programming, and the best systems rely on some form of statistical
learning, where parameters of hand-coded models are estimated from
large datasets of examples. However, the capabilities of
state-of-the-art machine learning systems are severely limited, and
only allow a small degree of adaptability of the machine's
functionality. For example, a speech recognition system will never be
able to perform speech translation by simply being instructed to do so--a human programmer is required to implement additional modules
manually.

\subsection{Long-term memory and com\-po\-si\-tio\-nal learning skills}

We see a special kind of long-term memory as the key component of the
intelligent machine. This long-term memory should be able to store
facts and algorithms corresponding to learned skills, making them accessible
on demand. In fact, even the ability to learn should be seen as a set
of skills that are stored in the memory. When the learning skills are
triggered by the current situation, they should compose new persistent
structures in the memory from the existing ones. Thus, the machine
should have the capacity to extend itself.

Without being able to store previously learned facts and skills, the
machine could not deal with rather trivial assignments, such as
recalling the solution to a task that has been encountered
before. Moreover, it is often the case that the solution to
a new task is related to that of earlier tasks. Consider for example
the following sequence of tasks in our simulated environment:

\begin{itemize}
\item find and pick an apple;
\item bring the apple back home;
\item find two apples;
\item find one apple and two bananas and bring them home.
\end{itemize}

Skills required to solve these tasks include:
\begin{itemize}
\item the ability to search around the current location;
\item the ability to pick things;
\item the ability to remember the location of home and return to it;
\item the ability to understand what \emph{one} and \emph{two} mean;
\item the ability to combine the previous skills (and more) to deal
  with different requests.
\end{itemize}

The first four abilities correspond to simple facts or skills to be
stored in memory: a sequence of symbols denoting something, the steps
needed to perform a certain action, etc. The last ability is an
example of a compositional \emph{learning skill}, with the capability
of producing new structures by composing together known facts and
skills. Thanks to such learning skills, the machine will be able to
combine several existing abilities to create a new one, often on the
fly. In this way, a well-functioning intelligent machine will not need
a myriad of training examples whenever it faces a slightly new
request, but it could succeed given a single example of the new
functionality. For example, when the Teacher asks the Learner to find one
apple and two bananas and bring them home, if the Learner already
understands all the individual abilities involved, it can retrieve the
relevant compositional learning skill to put together a plan and
execute it step by step. The Teacher may even call the new skill
generated in this way \emph{prepare breakfast}, and refer to it later
as such. Understanding this new concept should not require any further
training of the Learner, and the latter should simply store the new skill
together with its label in its long-term memory.

As we have seen in the previous examples, the Learner can continue
extending its knowledge of words, commands and skills in a completely unsupervised
way once it manages to acquire skills that allow it to compose structures
in its long-term memory. It may be that discovering the basic
learning skills, something we usually take for granted, is much
more intricate than it seems to us. But once we will be able to build a machine which
can effectively construct itself based on the incoming signals --even when
no explicit supervision in the form of rewards is given, as discussed above-- we
should be much closer to the development of intelligent machines.

\subsection{Computational properties of intelligent machines}
\label{sec:computational}

Another aspect of the intelligent machine that deserves discussion
is the computational model that the machine will be based on.  We are
convinced that such model should be unrestricted, that is, able to
represent any pattern in the data.  Humans can think of and talk about
algorithms without obvious limitations (although, to apply them, they
might need to rely on  external supports, such as paper and
pencil). A useful intelligent machine should be able to handle such
algorithms as well.

A more precise formulation of our claim in the context of the theory of computation
is that the intelligent machine needs to be based on a Turing-complete
computational model. That is, it has to be able to represent any
algorithm in fixed length, just like the Turing machine (the very fact
that humans can describe Turing-complete systems shows that they
are, in practical terms, Turing-complete: it is irrelevant, for our
purposes, whether human online processing capabilities are
strictly Turing-complete--what matters is that their reasoning skills,
at least when aided by external supports, are). Note that there
are many Turing-complete computational systems, and Turing machines in
particular are a lot less efficient than some alternatives, e.g.,
Random Access Machines. Thus, we are not interested in building the
intelligent machine around the concept of the Turing machine; we just
aim to use a computational model that does not have obvious
limitations in ability to represent patterns.

A system that is weaker than Turing-complete cannot represent certain
patterns in the data efficiently, which in turn means it cannot truly
learn them in a general sense. However, it is possible to memorize
such complex patterns up to some finite level of complexity.  Thus,
even a computationally restricted system may appear to work as
intended up to some level of accuracy, given that a sufficient number
of training examples is provided.

For example, we may consider a sequence repetition problem. The
machine is supposed to remember a sequence of symbols and reproduce it
later.  Further, let's assume the machine is based on a model with the
representational power of finite state machines. Such system is not
capable to represent the concept of storing and reproducing a
sequence. However, it may appear to do so if we design our experiment
imperfectly. Assume there is a significant overlap between what the
machine sees as training data, and the test data we use to evaluate
performance of the machine. A trivial machine that can function as a
look-up table may appear to work, simply by storing and recalling the
training examples. With an infinite number of  training examples, a
look-up-table-based machine would appear to learn any regularity. It
will work indistinguishably from a machine that can truly represent
the concept of repetition; however, it will need to have  infinite
size. Clearly, such memorization-based system will not perform well in
our setting, as we aim to test the Learner's ability to generalize
from a few examples.

Since there are many Turing-complete computational systems, one may
wonder which one should be preferred as the basis for machine
intelligence. We cannot answer this question yet, however we
hypothesize that the most natural choice would be a system that
performs computation in a parallel way, using elementary units that
can grow in number based on the task at hand. The growing property is
necessary to support the long-term memory, if we assume that the basic
units themselves are finite. An example of an existing computational
system with many of the desired properties is the cellular automaton
of \cite{von1966theory}. We might also be inspired by
 string rewriting systems, for example some versions
of the L-systems \citep{prusinkiewicz2012algorithmic}.

An apparent alternative would be to use a
non-growing model with immensely large capacity. There is however an
important difference. In a growing model, the new cells can be
connected to those that spawned them, so that the model is naturally
able to develop a meaningful topological structure based on functional
connectivity. We conjecture that such structure would in itself
contribute to learning in a crucial way.  On the other hand, it is not
clear if such topological structure can arise in a large-capacity
unstructured model. 
Interestingly, some of the more effective machine-learning models
available today, such as recurrent and convolutional neural networks,
are characterized by (manually constrained) network topologies that
are well-suited to the domains they are applied
to. 

\section{Related ideas}
\label{sec:related}


We owe, of course, a large debt to the seminal work of
\cite{Turing:1950}. Note that, while Turing's paper is most often
cited for the ``imitation game'', there are other very
interesting ideas in it, worthy of more attention from curious
readers, especially in the last section on learning machines. Turing
thought that a good way to construct a machine capable of passing his
famous test would be to develop a \emph{child machine}, and teach it
further skills through various communication channels.  These would
include sparse rewards shaping the behavior of the child machine, and
other information-rich channels such as language input from a teacher
and sensory information.

We share Turing's goal of developing a child machine capable of
independent communication through natural language, and we also stress
the importance of sparse rewards. The main distinction between his and
our vision is that Turing assumed that the child machine would be
largely programmed (he gives an estimate of sixty programmers working
on it for fifty years). We rather think of starting with a machine
only endowed with very elementary skills, and focus on the capability
to learn as the fundamental ability that needs to be developed. This
further assumes educating the machine at first in
a simulated environment where an artificial teacher will train it, as we outlined
in our roadmap. We also diverge with respect to the imitation game,
since the purpose of our intelligent machine is not to fool human
judges into believing it is actually a real person. Instead, we aim to
develop a machine that can perform a similar set of tasks to those a human
can do by using a computer, an Internet connection and the ability to
communicate.

There has been a recent revival of interest in tasks measuring
computational intelligence, spurred by the empirical advances of
powerful machine-learning architectures such as multi-layered neural
networks \citep{LeCun:etal:2015}, and by the patent inadequacy of the
classic version of Turing test \citep{Wikipedia:TuringTest:2015}. For
example, \cite{Levesque:etal:2012} propose to test systems on their
ability to resolve coreferential ambiguities (\emph{The trophy would
  not fit in the brown suitcase because it was too big\ldots What was
  too big?}).  \cite{Geman:etal:2015} propose a ``visual'' Turing test
in which a computational system is asked to answer a set of
increasingly specific questions about objects, attributes and
relations in a picture (\emph{Is there a person in the blue region? Is
  the person carrying something? Is the person interacting with any
  other object?}). Similar initiatives differ from ours in that they
focus on a specific set of skills (coreference, image parsing) rather
than testing if an agent can learn new skills. Moreover, these are
traditional evaluation benchmarks, unlike the hybrid
learning/evaluation ecosystem we are proposing.

The idea of developing an AI living in a controlled synthetic
environment and interacting with other agents through natural language
is quite old. The Blocks World of \cite{Winograd:1971} is probably the
most important example of early research in this vein. The approach
was later abandoned, when it became clear that the agents developed
within this framework did not scale up to real-world challenges
\citep[see, e.g.,][]{Morelli:etal:1992}. The knowledge encoded in the
systems tested by these early simulations was manually programmed by
their creators, since they had very limited learning
capabilities. Consequently, scaling up to the real world implied
manual coding of all the knowledge necessary to cope with it, and this
proved infeasible. Our simulation is instead aiming at systems that
encode very little prior knowledge and have strong capabilities to
learn from data. Importantly, our plan is not to try to manually
program all possible scripts our system might encounter later, as in
some of the classic AI systems. We plan to program only the initial
environment, in order to kickstart the machine's ability to learn and
adapt to different problems and scenarios. After the simulated
environment is mastered, scaling up the functionality of our Learner
will not require further manual work on scripting new situations, but
will rather focus on integrating real world inputs, such as those
coming from human users. The toy world itself is already designed to
feature novel tasks of increasing complexity, explicitly testing the
abilities of systems to autonomously scale up.

Still, we should not underestimate the drawbacks of synthetic
simulations. The tasks in our environment might directly address
some challenging points in the development of AI, such as learning with very weak supervision,
being able to form a structured long-term memory, and the ability of the
child machine to grow in size and complexity when encountering new problems.
However, simulating the real world can only bring us so far, and we might end up
overestimating the importance of some arbitrary phenomena at the
expense of others, that might turn out to be more common in natural
settings. It may be important to bring reality into the
picture relatively soon. Our toy world should let the intelligent machine
develop to the point at which it is able to learn from and cooperate with
actual humans. Interaction with real-life humans will then naturally
lead the machine to deal with real-world problems. The issue of when
exactly a machine trained in our controlled synthetic environment is
ready to go out in the human world is open, and it should be explored
empirically. However, at the same time, we believe that having
the machine interact with humans before it can deal with basic problems
in the controlled environment would be pointless, and possibly even
strongly misleading. 

Our intelligent machine shares some of its desired functionalities
with the current generation of automated personal assistants such as
Apple's Siri ad Microsoft's Cortana. However, these are heavily
engineered systems that aim to provide a natural language interface
for human users to perform a varied but fixed set of tasks (similar
considerations also apply to artificial human companions and digital
pets such as Tamagotchi, see
\citealp{Wikipedia:ArtificialHumanCompanion:2015}). Such systems can
be developed by defining the most frequent use cases, choosing those
that can be solved with the current technology (e.g., book an air
ticket, look at the weather forecast and set the alarm clock for
tomorrow's morning), and implementing specific solutions for each such
use case. Our intelligent machine is not intended to handle just a
fixed set of tasks. As exemplified by the example in Section
\ref{sec:advanced-examples}, the machine should be capable to learn
efficiently how to perform tasks such as those currently handled by
personal assistants, and more, just from interaction with the human
user (without a programmer or machine learning expert in the loop).

Architectures for software agents, and more specifically
\emph{intelligent} agents, are widely studied in AI and related fields
\citep{Nwana:1996,Russell:Norvig:2009}. We cannot review this ample
literature here, in order to position our proposal precisely with
respect to it. We simply remark that we are not aware of other
architectures that are as centered on learning and communication
as ours. Interaction plays a central role in the study of
multiagent systems \citep{Shoham:LeytonBrown:2009}. However, the
emphasis in this research tradition is on how conflict resolution and
distributed problem solving evolve in typically large groups of
simple, mostly scripted agents. For example, traffic modeling is a
classic application scenario for multiagent systems. This is very
different from our emphasis on linguistic interaction for the purposes
of training a single agent that should become independently capable of
very complex behaviours.

\cite{Tenenbaum:2015}, like us, emphasizes the need to focus on basic
abilities that form the core of intelligence. However, he takes naive
physics problems as the starting point, and discusses specific classes
of probabilistic models, rather than proposing a general learning
scenario. There are also some similarities between our proposal and
the research program of Luc Steels
\citep[e.g.,][]{Steels:2003,Steels:2005}, who lets robots evolve
vocabularies and grammatical constructions through interaction in a
situated environment. However, on the one hand his agents are actual
robots subject to the practical hardware limitations imposed by the
need to navigate a complex natural environment from the start; on the
other, the focus of the simulations is narrowly on language
acquisition, with no further aim to develop broadly intelligent
agents.

We have several points of contact with the semantic parsing
literature, such as navigation tasks in an artificial world
\citep{MacMahon:etal:2006} and reward-based learning from natural
language instructions
\citep{Chen:Mooney:2011,Artzi:Zettlemoyer:2013}. The agents developed
in this area can perform tasks, such as learning to execute
instructions in natural environments by interacting with humans
\citep{Thomason:etal:2015}, or improving performance on real-life
video-games by consulting the instruction manual
\citep{Branavan:etal:2012}, that we would want our intelligent
machines to also be able to carry out. However, current
semantic-parsing-based systems achieve these impressive feats by
exploiting architectures tuned to the specific tasks at hand, and they
rely on a fair amount of hard-wired expert knowledge, in particular
about language structures (although recent work is moving towards a
more knowledge-lean direction, see for example
\citealp{Narasimhan:etal:2015}, who train a neural network to play
text-based adventure games using only text descriptions as input and
game reward as signal). Our framework is meant to encourage the
development of systems that should eventually be able to perform
similar tasks, but getting there incrementally, starting with almost
no prior knowledge and first learning from their environment a set of
simpler skills, and how to creatively merge them to tackle more
ambitious goals.

The last twenty years have witnessed several related proposals on
learning to learn \citep{Thrun:Pratt:1997}, lifelong learning
\citep{Silver:etal:2013} and continual learning
\citep{Ring:1997}. Much of this work is theoretical in nature and
focuses on algorithms rather than on empirical challenges for the
proposed models. Still, the general ideas being pursued are in line
with our program. \cite{Ring:1997}, in particular, defines a
continual-learning agent whose experiences ``occur sequentially, and
what it learns at one time step while solving one task, it can use
later, perhaps to solve a completely different task.'' Ring's
desiderata for the continual learner are remarkably in line with
ours. It is ``an autonomous agent. It senses, takes actions, and
responds to the rewards in its environment. It learns behaviors and
skills while solving its tasks. It learns incrementally. There is no
fixed training set; learning occurs at every time step; and the skills
the agent learns now can be used later.  It learns
hierarchically. Skills it learns now can be built upon and modified
later.  It is a black box. The internals of the agent need not be
understood or manipulated. All of the agent's behaviors are developed
through training, not through direct manipulation. Its only interface
to the world is through its senses, actions, and rewards.  It has no
ultimate, final task. What the agent learns now may or may not be
useful later, depending on what tasks come next.'' Our program is
definitely in the same spirit, with an extra emphasis on
interaction. 

\cite{Mitchell:etal:2015} discuss NELL, the most fully realized
concrete implementation of a lifelong learning architecture. NELL is
an agent that has been ``reading the Web'' for several years to
extract a large knowledge base. Emphasis is on the never-ending nature
of the involved tasks, on their incremental refinement based on what
NELL has learned, and on sharing information across tasks. In this
latter respect, this project is close to multi-task learning
\citep{Ando:Zhang:2005,Caruana:1997,Collobert:etal:2011}, that focuses
on the idea of parameter sharing across tasks. It is likely that a
successful learner in our framework will exploit similar strategies,
but our current focus lies on defining the tasks, rather than on how
to pursue them.

\cite{Bengio:etal:2009} propose the related idea of curriculum
learning, whereby training data for a single task are ordered
according to a difficulty criterion, in the hope that this will lead
to better learning. This is motivated by the observation that humans
learn incrementally when developing complex skills, an idea that has
also previously been studied in the context of recurrent neural
network training by \cite{Elman:1993}. %
The principle of incremental learning is
also central to our proposal. However, the fundamental aspect for us
is not a strict ordering of the training data for a specific task, but
incrementality in the \emph{skills} that the intelligent machine
should develop. This sort of incrementality should in turn be boosted
by designing separate tasks with a compositional structure, such that
the skills acquired from the simpler tasks will help to solve the more
advanced ones more efficiently.

The idea of incremental learning, motivated by the same considerations
as in the papers we just mentioned, also appears in
\cite{solomonoff2002progress}, a work which has much earlier roots in
research on program induction
\citep{solomonoff1964formal,solomonoff1997discovery,schmidhuber2004optimal}. Within
this tradition, \cite{Schmidhuber:2015} reviews a large literature and
presents some general ideas on learning that might inspire our search
for novel algorithms. Genetic programming \citep{Poli:etal:2008} also
focuses on the reuse of previously found sub-solutions, speeding up
the search procedure in this way. Our proposal is also related to that
of \cite{Bottou:2014}, in its vision of compositional machine
learning, although he only considers composition in limited domains,
such as sentence and image processing.

We share many ideas with the reinforcement learning
framework \citep{Sutton:Barto:1998}.  In reinforcement learning, the
agent chooses actions in an environment in order to maximize some
cumulative reward over time. Reinforcement learning is particularly
popular for problems where the agent can collect information only by
interacting with the environment.  Given how broad this definition is,
our framework could be considered as a particular instance of it.  Our
proposal is however markedly different from standard reinforcement
learning work \citep{kaelbling:etal:1996} in several
respects. Specifically, we emphasize language-mediated, interactive
communication, we focus on incremental strategies that encourage agents to solve
tasks by reusing previously learned knowledge and we aim to limit the
number of trials an agent gets in order to accomplish a certain goal.

\cite{Mnih:etal:2015} recently presented a single neural network
architecture capable of learning a set of classic Atari games using
only pixels and game scores as input (see also the related idea of
``general game playing'', e.g., \citealp{Genesereth:etal:2005}). We
pursue a similar goal of learning from a low-level input stream and
reward. However, unlike these authors, we do not aim for a single
architecture that can, disjointly, learn an array of separate tasks,
but for one that can incrementally build on skills learned on previous
tasks to perform more complex ones. Moreover, together with reward, we
emphasize linguistic interaction as a fundamental mean to foster skill
extension. \cite{Sukhbaatar:etal:2015b} introduce a sandbox to design
games with the explicit purpose to train computational agents in
planning and reasoning tasks. Moreover, they stress a curriculum
strategy to foster learning (making the agent progress through
increasingly more difficult versions of the game). Their general
program is aligned with ours, and the sandbox might be useful to
develop our environment. However, they do not share our emphasis on
communication and interaction, and their approach to incremental
learning is based on increasingly more difficult versions of the same
task (e.g., increasing the number of obstacles), rather than on
defining progressively more complex tasks, such that solving 
the later ones requires composing solutions to earlier ones, as we are
proposing. Furthermore, the tasks currently considered within the
sandbox do not seem to be challenging enough to require new learning approaches, and may be
solvable with current techniques or minor modifications thereof.

\cite{mikolov2013incremental} originally discussed a preliminary
version of the incremental task-based approach we are more fully
outlying here. %
In a similar spirit, \cite{Weston:etal:2015b} present a set of
question answering tasks based on synthetically generated
stories. They also want to foster non-incremental progress in AI, but
their approach differs from ours in several crucial aspects. Again, there is
no notion of interactive, language-mediated learning, a classic
train/test split is enforced, and the tasks are not designed to
encourage compositional skill learning (although Weston and
colleagues do emphasize that the same system should be used for all
tasks). Finally, the evaluation metric is notably different from ours - while
we aim to minimize the number of trials it takes for the machine to master
the tasks, their goal is to have a good performance on held out data. This could
be a serious drawback for works that involve artificial tasks, as in
our view the goal should be to develop a machine that can learn as fast as possible,
to have any hope to scale up and be able to generalize in more complex
scenarios.

%



One could think of solving sequence-manipulation problems such as
those  constituting the basis of our learning routine with
relatively small extensions of established machine learning techniques
\citep{Graves:etal:2014,Grefenstette:etal:2015,Joulin:Mikolov:2015}. 
As discussed in the previous section, for simple tasks that involve
only a small, finite number of configurations, one could be apparently
successful even just by using a look-up table storing all possible
combinations of inputs and outputs. The above mentioned works, that
aim to learn algorithms from data, also add a long-term memory (e.g.,
a set of stacks), but they use it to store the data only, not the
learned algorithms. Thus, such approaches fail to generalize in
environments where solutions to new tasks are composed of already
learned algorithms.

Similar criticism holds for approaches that try to learn certain
algorithms by using an architecture with a strong prior towards their
discovery, but not general enough to represent even small
modifications. To give an example from our own work: a recurrent
neural network augmented with a stack structure can form a simple kind
of long-term memory and learn to memorize and repeat sequences in the
reversed order, but not in the original one
\citep{Joulin:Mikolov:2015}. We expect a valid solution to the
algorithmic learning challenge to utilize a small number of training
examples, and to learn tasks that are closely related at an increasing
speed, i.e., to require less and less examples to master new skills
that are related to what is already known.  We are not aware of any
current technique addressing these issues, which were the very reason
why algorithmic tasks were originally proposed by
\cite{mikolov2013incremental}. We hope that this paper will motivate
the design of the genuinely novel methods we need in order to develop
intelligent machines.

\section{Conclusion}
\label{sec:conclusion}

We defined basic desiderata for an intelligent machine, stressing
learning and communication as its fundamental abilities. Contrary to
common practice in current machine learning, where the focus is on
modeling single skills in isolation, we believe that all aspects of
intelligence should be holistically addressed within a single system.

We proposed a simulated environment that requires the intelligent
machine to acquire new facts and skills through communication.
In this environment, the machine must learn to
perform increasingly more ambitious tasks, being naturally induced to
develop complex linguistic and reasoning abilities.

We also presented some conjectures on the properties of the
computational system that the intelligent machine may be based on.
These include learning of algorithmic patterns from a few examples without strong
supervision, and development
of a long-term memory to store both data and learned skills. We
tried to put this in contrast with currently accepted paradigms in
machine learning, to show that 
current methods are far from adequate, and we must
strive to develop non-incrementally novel techniques.

This roadmap constitutes only the beginning of a long journey towards
AI, and we hope other researchers will be joining it in pursuing the
goals it outlined. 

\section*{Acknowledgments}

We thank L\'{e}on Bottou, Yann LeCun, Gabriel Synnaeve, Arthur Szlam,
Nicolas Usunier, Laurens van der Maaten, Wojciech Zaremba and others
from the Facebook AI Research team, as well as Gemma Boleda, Katrin
Erk, Germ\'{a}n Kruszewski, Angeliki Lazaridou, Louise McNally,
Hinrich Sch\"{u}tze and Roberto Zamparelli for many stimulating
discussions. An early version of this proposal has been discussed in
several research groups since 2013 under the name \emph{Incremental
  learning of algorithms} \citep{mikolov2013incremental}.

\bibliographystyle{apalike}
\bibliography{marco,armand,tomas}

\end{document}